\documentclass[10pt,twocolumn,letterpaper]{article}

\usepackage[pagenumbers]{cvpr} 

\usepackage[dvipsnames]{xcolor}

\definecolor{cvprblue}{rgb}{0.21,0.49,0.74}

\usepackage{array}
\usepackage{subcaption}
\usepackage{multirow,multicol}
\usepackage[flushleft]{threeparttable}
\usepackage{comment}
\usepackage{algorithmic}
\usepackage{booktabs}
\usepackage{bbm, dsfont}
\usepackage[font=small,labelfont=bf]{caption}

\usepackage{amssymb}
\usepackage{pifont}

\newcolumntype{x}[1]{>{\centering\arraybackslash}p{#1pt}}
\newcommand{\app}{\raise.17ex\hbox{$\scriptstyle\sim$}}

\newlength\savewidth

\makeatletter\renewcommand\paragraph{\@startsection{paragraph}{4}{\z@}
  {.5em \@plus1ex \@minus.2ex}{-.5em}{\normalfont\normalsize\bfseries}}\makeatother

\def\tablecite#1#{%
  \def\pretablecite{#1}%
  \tableciteaux}
\def\tableciteaux#1{%
  \textsuperscript{\expandafter\originalcite\pretablecite{#1}}%
}

\usepackage{graphicx}
\usepackage{enumitem}
\usepackage{wrapfig}
\usepackage{lipsum}
\usepackage{soul}

\usepackage{color, colortbl}

\usepackage{tabu}
\usepackage{xcolor}
\usepackage{nicematrix}

\definecolor{ForestGreen}{rgb}{0.13, 0.55, 0.13}
\definecolor{Green}{rgb}{0.0, 0.5, 0.0}
\definecolor{Blue}{rgb}{0.25, 0.42, 0.88}
\definecolor{green(munsell)}{rgb}{0.0, 0.66, 0.47}
\definecolor{green(ryb)}{rgb}{0.4, 0.69, 0.2}
\definecolor{green(pigment)}{rgb}{0.0, 0.65, 0.31}
\definecolor{citecolor}{HTML}{0071bc}
\definecolor{GrayXMark}{gray}{0.7}
\definecolor{DifferenceColor}{HTML}{af3235}
\definecolor{HighlightColor}{gray}{0.9}
\definecolor{OracleTextColor}{gray}{0.55}
\definecolor{Cerulean}{HTML}{00a2e3}

\newcolumntype{H}{>{\setbox0=\hbox\bgroup}c<{\egroup}@{}}
\newcolumntype{a}{>{\columncolor{HighlightColor}}c}
\newcolumntype{L}[1]{>{\centering\arraybackslash}m{#1}}

\usepackage{tabularx}
\usepackage[export]{adjustbox}
\usepackage{makecell}
\usepackage{ragged2e}

\usepackage{dblfloatfix}

\usepackage[pagebackref,breaklinks,colorlinks,citecolor=cvprblue]{hyperref}

\usepackage[capitalize]{cleveref}
\crefname{section}{\S}{\S\S}
\crefname{subsection}{\S}{\S\S}
\crefformat{table}{Table~#2#1#3}
\crefformat{figure}{Figure~#2#1#3}
\crefformat{equation}{Eq~#2#1#3}

\def\paperID{240} 
\def\confName{CVPR}
\def\confYear{2024}

\title{PointRWKV: Efficient RWKV-Like Model for Hierarchical Point Cloud Learning}

\author{
  Qingdong He$^1$\footnotemark[1]
  ~~ Jiangning Zhang$^1$\thanks{Equal contributions.}
  ~~ Jinlong Peng$^1$ 
  ~~ Haoyang He$^2$\\
  ~~  Xiangtai Li $^3$
  ~~ Yabiao Wang$^1$
  ~~ Chengjie Wang$^1$ 
   \\ \vspace{4pt}
   \normalsize $^1$Youtu Lab, Tencent ~~ $^2$Zhejiang University ~~ $^3$Nanyang Technological University\\
}

\begin{document}
\maketitle







\begin{abstract}
Transformers have revolutionized the point cloud learning task, but the quadratic complexity hinders its extension to long sequences. 
This puts a burden on limited computational resources. 
The recent advent of RWKV, a fresh breed of deep sequence models, has shown immense potential for sequence modeling in NLP tasks. 
In this work, we present PointRWKV, a new model of linear complexity derived from the RWKV model in the NLP field with the necessary adaptation for 3D point cloud learning tasks. 
Specifically, taking the embedded point patches as input, we first propose to explore the global processing capabilities within PointRWKV blocks using modified multi-headed matrix-valued states and a dynamic attention recurrence mechanism. 
To extract local geometric features simultaneously, we design a parallel branch to encode the point cloud efficiently in a fixed radius near-neighbors graph with a graph stabilizer. 
Furthermore, we design PointRWKV as a multi-scale framework for hierarchical feature learning of 3D point clouds, facilitating various downstream tasks. 
Extensive experiments on different point cloud learning tasks show our proposed PointRWKV outperforms the transformer- and mamba-based counterparts, while significantly saving about 42\% FLOPs, demonstrating the potential option for constructing foundational 3D models. 
Project page: \url{https://hithqd.github.io/projects/PointRWKV/}.
\end{abstract}

\section{Introduction}
\label{sec:intro}

Point cloud analysis is a fundamental vision task that has a wide range of applications, including autonomous driving~\cite{mao20223d}, virtual reality~\cite{park2008multiple}, and robotics~\cite{shafiullah2022clip}, \textit{etc}. Unlike 2D images, the intrinsic irregularity and sparsity of point clouds make it a challenging task to conduct point cloud representation learning. Balancing accuracy and complexity simultaneously remains an enduring problem.

\begin{figure}[t]
    \centering
\includegraphics[width=1\linewidth]{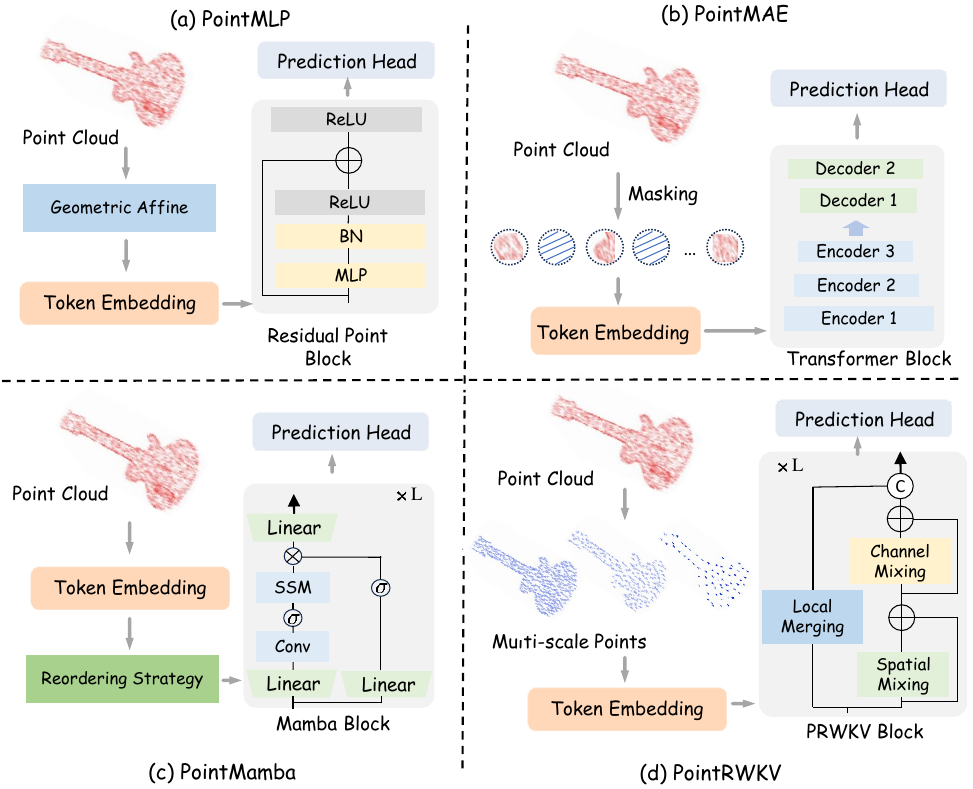}
    \caption{\textbf{Architecture comparison with different methods:} (a) MLP-based PointMLP~\cite{ma2022rethinking}, (b) transformer-based PointMAE~\cite{pang2022masked}, (c) mamba-based PointMamba~\cite{liang2024pointmamba} and (d) ours PointRWKV with linear complexity is capable of integrating the advantages of both global and local modeling, and multi-scale features endow it with more refined prediction accuracy.}
    \label{fig::motivation}
\end{figure}

\begin{figure}[t]
    \centering
\includegraphics[width=0.95\linewidth]{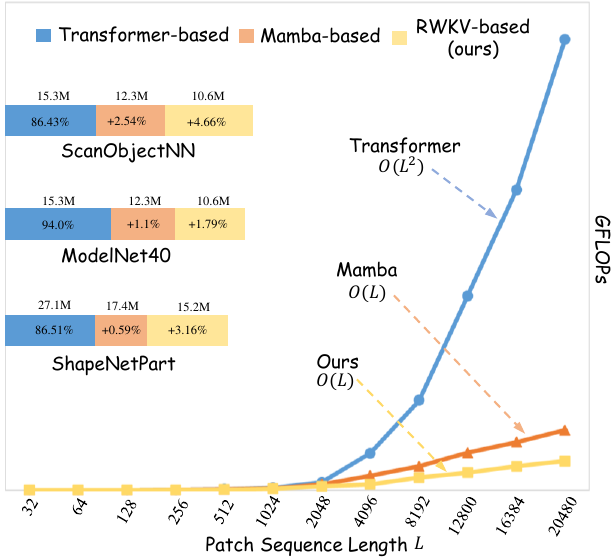}
    \caption{\textbf{Accuracy-speed tradeoff.} (Left) Overall accuracy acquired by different methods with relative parameters, (Right) FLOPs increase with sequence length.}
    \label{fig:acc_time}
\end{figure}

Pioneered by PointNet~\cite{qi2017pointnet} and PointNet++~\cite{qi2017pointnet++}, a series of works~\cite{atzmon2018point,li2018pointcnn,komarichev2019cnn,qian2022pointnext,thomas2019kpconv,wu2019pointconv,ma2022rethinking,wang2019dynamic} have emerged to follow the MLP architecture to extract geometric features (Figure~\ref{fig::motivation} (a)). 
Recently, transformers~\cite{dosovitskiy2020image,liu2021swin} have surfaced as a promising approach for point cloud analysis, demonstrating significant potential with a key component of the attention mechanism to effectively capture the relationships among a set of points. 
Focusing on enhancing transformers for point clouds, Point Transformers~\cite{zhao2021point,wu2022point,wu2023point} apply vector attention and enhance the performance and efficiency for different tasks. Further, by extending self-supervised learning mechanism, some works~\cite{yu2022point,pang2022masked,zhang2022point,qi2023contrast} have obtained decent performance. Among them, PointMAE~\cite{pang2022masked}, as depicted in Figure~\ref{fig::motivation} (b), introduces a standard Transformer architecture and a shifting mask tokens operation at a high ratio to reach better accuracy. 
However, directly applying exhaustive attention mechanisms to extended point tokens increases the demand for computational resources.
This is because of the quadratic complexity inherent in attention computations, impacting both computation and memory. 
Furthermore, the recent state space models (SSMs)~\cite{gu2023mamba,zhu2024vision,liu2024vmamba} introduce an alternative approach for point sequence modeling. Based on the structured SSMs, several works~\cite{liang2024pointmamba,zhang2024point,liu2024point,han2024mamba3d} have attempted to enable the model to causally capture the point cloud structure with consistent traverse serialization mechanism or reordering strategy, as illustrated in Figure~\ref{fig::motivation} (c).
Albeit effective, the inherent property of unidirectional modeling of the vanilla SSM hinders them from achieving superior performance for the unordered point cloud data. 
%
Considering this, one essential question arises: \emph{How to design a method that runs with linear complexity and achieves a high performance simultaneously?}

We notice the recent advance of RWKV~\cite{peng2023rwkv} model within natural language processing (NLP), serving as a promising solution for attaining enhanced efficiency and handling extensive textual data. 
%
This paper pioneers the application of RWKV into the 3D point cloud learning domain, introducing PointRWKV, as illustrated in Figure~\ref{fig::motivation}(d).
PointRWKV maintains the core structure and benefits of RWKV while preserving the linear complexity and processing multi-scale point input. 
Significantly, it exhibits a reduced parameter count and computational requirement, rendering it aptly suitable for deep point cloud learning.

Specifically, PointRWKV utilizes RWKV-like architecture to model the global features of point clouds with a series of stacked PRWKV blocks. Considering the unique properties of the strong geometric and semantic dependence between local parts and the overall 3D shapes of point cloud data, we modify the architecture into multi-stage hierarchies for progressively encoding multi-scale features of point clouds akin to an asymmetric U-Net~\cite{ronneberger2015u}. 
%
Within each PRWKV block, we devise a parallel strategy for hierarchical point cloud feature representation, enabling local and global feature aggregation. 
For the original RWKV branch, we present the bidirectional quadratic expansion function with multi-headed matrix-valued states and adapt the original causal attention mechanism into a dynamic bidirectional global attention mechanism. 
The former can broaden the semantic scope of individual tokens, while the latter facilitates the computation of global attention within linear computation complexity. 
For another branch, the graph stabilizer mechanism is proposed to conduct local graph-based merging for learning the compact representation of the point cloud. 
These designs boost the model's capacity while maintaining scalability and stability. 
In this way, our PointRWKV inherits the efficiency of RWKV in processing global information and sparse inputs while also being capable of modeling the local concept of point clouds.

By employing multi-scale pre-training, our PointRWKV can encode point clouds from local-to-global hierarchies, learning robust 3D representations and demonstrating exceptional transfer capabilities. 
We conduct extensive experiments on various point cloud learning tasks (\textit{e.g.}, classification, part segmentation, and few-shot learning) to demonstrate the effectiveness of our method. As shown in Figure~\ref{fig:acc_time}, after self-supervised pre-training on ShapeNet~\cite{chang2015shapenet}, PointRWKV achieves 93.63$\%$ (+4.66$\%$) overall accuracy on the ScanObjectNN~\cite{uy2019revisiting} and 96.89$\%$ (+1.79$\%$) accuracy on ModelNet40~\cite{wu20153d} for shape classification, 90.26$\%$ (+3.16$\%$) instance mIoU on ShapeNetPart~\cite{yi2016scalable} for part segmentation, setting new state-of-the-art (SoTA) among pre-trained models. Meanwhile, PointRWKV reduces 13$\%$ in parameters and 42$\%$ in FLOPs compared to the transformer- and mamba-based counterparts, demonstrating the potential of RWKV in the 3D vision tasks. 
Our contributions are summarized as follows:
\begin{itemize}
\item We propose PointRWKV, which innovatively applies the RWKV framework to address point cloud learning tasks. This approach reaches a better balance among the parameters, computation complexity, and efficiency, making it an alternative for 3D vision tasks.
\item To learn the hierarchical representations of point clouds, we introduce the multi-stage hierarchies architecture to encode multi-scale point features. Within each block in this architecture, we design the parallel strategy for hierarchical local and global feature aggregation, equipped with the bidirectional quadratic expansion, the dynamic bidirectional global attention mechanism, and the graph stabilizer mechanism.
\item Our PointRWKV exhibits state-of-the-art performance on various downstream tasks, outperforming  transformer- and mamba-based counterparts in both efficiency and computation complexity, which indicates our approach to be a powerful representation learner for 3D point clouds.
\end{itemize}

\section{Related Work}
\label{sec:related_work}

\noindent\textbf{MLP/CNN-based Models For Point Cloud.} With the progressive evolution of deep neural networks, point cloud learning has gained increasing interest, leading to the emergence of numerous deep architectures and models in recent years~\cite{qi2017pointnet,huang2022adaptive,deng2023pointvector,ma2022rethinking,huang20223qnet,huang2023learning}. 
PointNet~\cite{qi2017pointnet} and PointNet++~\cite{qi2017pointnet++} are the two pioneering approaches that use MLPs directly to process point clouds. 
Following these, a series of works~\cite{atzmon2018point,choy20194d,deng2023pointvector,li2018pointcnn,zhao2019pointweb,komarichev2019cnn,qian2022pointnext,thomas2019kpconv,wu2019pointconv,ma2022rethinking} attempt to design various deep architectures to better capture local and global geometric features. 
Meanwhile, several works~\cite{he2022svga,landrieu2018large,wang2019dynamic} explore to utilize graph neural networks to construct the 3D geometric relationships between the points.

\noindent\textbf{Transformer-based Models For Point Cloud.} Equipped with attention mechanism, transformer~\cite{vaswani2017attention} has evolved not only in NLP~\cite{devlin2018bert,radford2018improving} but also in 2D~\cite{dosovitskiy2020image,liu2021swin} and 3D vision~\cite{misra2021end,mao2021voxel,hatamizadeh2022unetr}, including the point cloud learning~\cite{guo2021pct,zhao2021point,yu2021pointr,park2022fast}. The Point Transformer series~\cite{zhao2021point,wu2022point,wu2023point} are pioneering in introducing self-attention layers specifically tailored for point clouds. Significant advancements are marked by the implementation of grouped vector attention and partition-based pooling techniques for enhanced point cloud analysis. 
Recently, a surge of research~\cite{yu2022point,pang2022masked,zhang2022point} efforts have been directed towards enhancing point cloud data representation through self-supervised learning, particularly by leveraging masked point modeling. This self-supervised approach enables models to learn the intrinsic structure and features of point clouds in the absence of explicit labels, fostering a novel paradigm in the field. 

\noindent\textbf{Mamba-based Models.} The Mamba model, a deep learning architecture grounded in state-space models (SSMs)~\cite{gu2023mamba,mehta2022long,fu2022hungry,smith2022simplified}, exhibits enhanced processing speed and scalability by enabling selective memory retention or discarding of inputs. 
In recent studies, the Mamba model has demonstrated its efficacy across a diverse range of applications, such as NLP~\cite{pioro2024moe,anthony2024blackmamba}, 2D image analysis~\cite{liu2024vmamba,he2024mambaad}, and video processing~\cite{yang2024vivim,yang2024vivim}, showcasing its versatility and adaptability. Recent researches~\cite{liang2024pointmamba,zhang2024point,liu2024point,han2024mamba3d} have successfully employed the Mamba model in 3D point cloud tasks, where point cloud data is represented as state vectors, allowing the model to capture the dynamic characteristics of the data, resulting in a commendable performance.

\begin{figure*}[ht]
\centering
\includegraphics[width=0.95\linewidth]{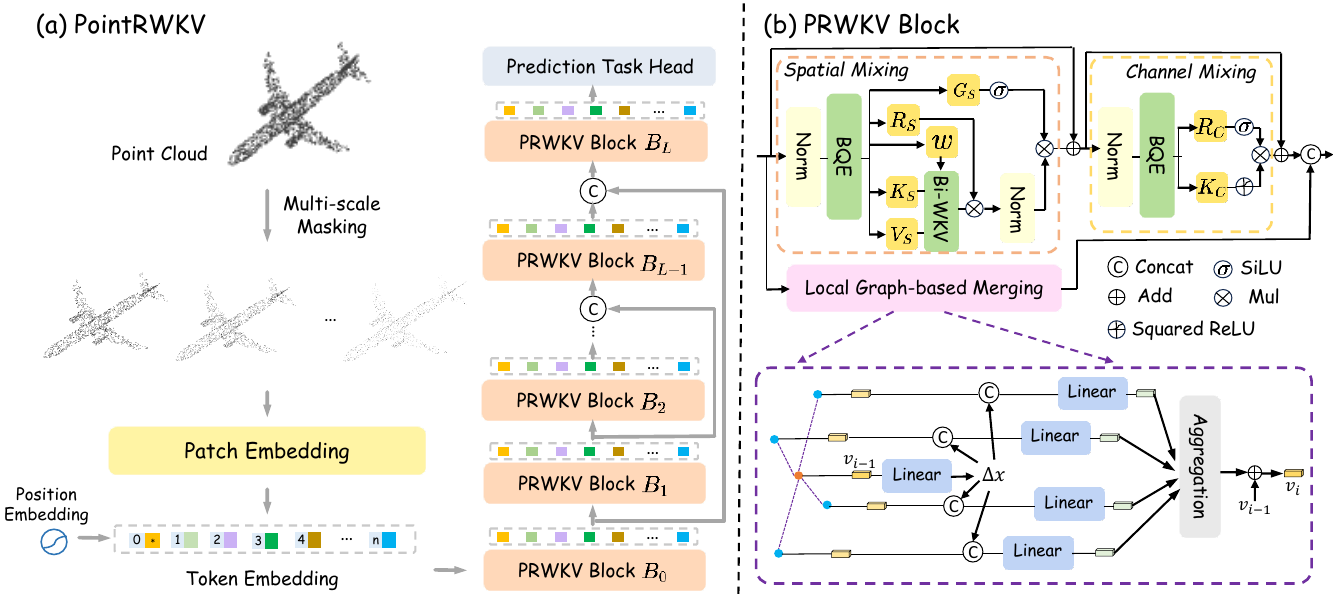}
\caption{\textbf{Overview of the proposed PointRWKV}, which employs a hierarchical architecture to encode multi-scale point cloud features. The whole framework is composed of a series of PRWKV blocks which include the integrative feature modulation branch and the local graph-based merging branch to form the parallel feature learning strategy.}
\label{fig::pointrwkv}
\end{figure*}

\noindent
\textbf{Receptance Weighted Key Value (RWKV) Models.} 
Transformer's memory and computational complexity scale quadratically with sequence length, while RNNs exhibit linear growth in these requirements, albeit limited by parallelization and scalability. Recently, the innovative Receptance Weighted Key Value (RWKV)~\cite{peng2023rwkv,peng2024eagle} model is introduced, seamlessly combining Transformer's efficient parallel training with RNN's efficient inference. Empirical results demonstrate that RWKV exhibits linear time complexity and holds promise for outperforming Transformers in long-sequence reasoning tasks. Vision-RWKV~\cite{duan2024vision} is an advanced adaptation of the RWKV mechanism tailored for visual tasks, leveraging its linear computational complexity to deliver exceptional efficiency in processing high-resolution images. The following works~\cite{yuan2024rwkvsam,yang2024restorerwkv} also explore RWKV on dense prediction tasks.
In this paper, we develop RWKV to exploit its modeling capacity and linear computational efficiency for multi-class unsupervised, making it an attractive solution for 3D point cloud learning applications.

\section{Method}
The overall pipeline of PointRWKV is shown in Figure~\ref{fig::pointrwkv}(a), where we encode the point cloud by a hierarchical network architecture. 
Given an input point cloud, we first employ the multi-scale masking strategy~\cite{zhang2022point} to sample different point numbers at various scales. 
Then a lightweight PointNet~\cite{qi2017pointnet} is applied to embed point patches and generate token embeddings. 
These point tokens are consumed by the block-stacked encoder, namely the PRWKV block, where each block consists of two parallel branches for hierarchical local and global feature aggregation. 

\subsection{Hierarchical Point Cloud Learning}
Taking an input point cloud $\mathbf{P} \in \mathbb{R}^{N \times 3}$, we apply the multi-scale masking to obtain the $M$ scales point clouds. Starting with the point cloud as the initial $M$-th scale, the process involves iterative downsampling and grouping using Furthest Point Sampling (FPS) and $k$ Nearest-Neighbour ($k$-NN). This results in a series of scales, each with a decreasing number of points. Then, a significant proportion of seed points at the final scale are randomly masked, leaving a subset of visible points. These visible points are then back-projected across all scales to ensure consistency. This involves retrieving the $k$ nearest neighbors from the subsequent scale's indices to serve as the visible positions for the current scale, with the remaining positions masked. The process yields visible and masked positions for all scales, providing a comprehensive multi-scale representation of the original point cloud. Finally, we employ a mini-PointNet to extract features for each local patch of different scales and add the positional encoding to get the final sequence, serving as the initial token embeddings $\mathbf{E_0} \in \mathbb{R}^{T \times C}$.

\noindent\textbf{PRWKV Block.}
After obtaining the token embeddings, we feed them into the encoder, containing a series of PRWKV blocks with long skip connections between shallow and deep layers. Specifically, for each PRWKV block, as shown in \ref{fig::pointrwkv}(b), the two parallel branches processing strategy is employed to aggregate local and global features. The upper is the \emph{integrative feature modulation (IFM)} flow with spatial-mixing and channel-mixing, and the below is the \emph{local graph-based merging (LGM)}. Finally, the concatenation of the two branches is used as the output of each block.

\subsection{Integrative Feature Modulation (IFM)}\label{sec:IFM}
The integrative feature modulation branch consists of a spatial-mixing module and a channel-mixing module. The spatial-mixing module operates as an attention mechanism, executing global attention computations of linear complexity, whereas the channel-mixing module functions as a feed-forward network (FFN), facilitating feature fusion along the channel dimension.

\noindent\textbf{Spatial-Mixing.}
After one pre-LayerNorm, the tokens are first shifted by the bidirectional quadratic expansion (BQE) function and then fed into four parallel linear layers to obtain the multi-headed vectors $R_S$, $K_S$,$V_S$, $G_S$:
\begin{equation}
\sqcap _S = BQE_\sqcap (X){\mathbf{W_\sqcap^S }}, \sqcap \in \{R, K, V, G \}, 
\end{equation}
where BQE is formulated as :
\begin{equation}
\begin{aligned}
&  BQE_\mho (X) = X + (1 - \mu_\mho )X^\star  \\
& X^\star = \operatorname{\mathbf{Concat}}(X_1, X_2, X_3, X_4), 
\end{aligned}
\end{equation}
where $\mho \in \{R, K, V, G, w\}$, $\mu_\mho$ is the learnable vector and the $X^\star$ is the slicing vector of $X$, \emph{i.e.}, $X_1=[b-1,n,0:C/4]$, $X_2=[b+1,n,C/4:C/2]$, $X_3=[b,n-1,C/2:3C/4]$, $X_4=[b,n+1,3C/4:C]$, $b$ and $n$ are the dimension index of $X$. The BQE function empowers the attention mechanism to inherently focus on proximate tokens across diverse channels without requiring a considerable increment in FLOPs. This procedure also broadens the receptive field of each token, thereby significantly boosting the token's coverage in the ensuing layers. Further, a new time-varying decay $w$ is calculated by:
\begin{equation}
\begin{aligned}
& \nu(c) = \lambda_x + \tanh (c\mathbf{A_x})\mathbf{B_x} \\
& w_{\_}BQE (X) = X + (1 - \nu(BQE_w (X)) )X^\star \\
& d = \nu(w_{\_}BQE (X))\\
& w= \exp (-\exp (d), 
\end{aligned}
\end{equation}
where $\lambda_x$ is a trainable vector and $\mathbf{A_x}$, $\mathbf{B_x}$ are both trainable weight matrices.

Then, $K_S$ and $V_S$ are passed to calculate the global attention result $wkv$ with the new decay $w$. Here, we introduce the linear complexity bidirectional attention mechanism with two modifications: (1) the decay parameter varies independently, which is data-dependent in a dynamic manner, and (2) the upper limit of original RWKV attention is extended from the current token $t$ to the last token $T-1$ in the summation formula to ensure that all tokens are mutually visible in the calculation of each result. For the $t$-th token, the attention result is calculated by the following formula:
\begin{align}
wkv_t = diag(u_1, \dots, u_n)\cdot K_{S,t}^\mathsf{T}\cdot V_{S,t} \nonumber\\ 
        + \sum_{i=1 }^{T-1}diag(\varpi _{i,j}^1, \dots, \varpi _{i,j}^n) \cdot K_{S,i}\cdot V_{S,i}
\end{align}
where $u$ is a per-channel learned boost and $\varpi _{i,j} = \prod_{j=1 }^{i-1}w_j$ is a dynamic decay. The final probability output $O_S$ is :
\begin{equation}
\small
O_S = \operatorname{\mathbf{Concat}}(\sigma (G_S)\otimes LayerNorm(R_S\otimes wkv))\mathbf{W_O^S}
\end{equation}
where $\sigma$ is the SiLU activation function and $\otimes$ is the element-wise multiplication.

\noindent\textbf{Channel-Mixing.}
The tokens from the spatial-mixing module are further passed into the channel-mixing module. Similarly, the pre-LayerNorm is utilized and $R_C$, $K_C$ are obtained after the BQE operation:
\begin{equation}
\cap _C= BQE_\cap (X){\mathbf{W_\cap ^C }}, \cap  \in \{R, K \}.
\end{equation}

Afterward, a linear projection and a gate mechanism are performed, respectively. And the final output $O_C$ is formulated as:
\begin{equation}
O_C = (\sigma (R_C)\otimes SquaredReLU(K_C)\mathbf{W_V^C})\mathbf{W_O^C}.
\end{equation}

\subsection{Local Graph-based Merging (LGM)}
\label{LGM}

Local geometric features have been proven to be crucial for point cloud feature learning, but RWKV’s global receptive field cannot comprehensively capture local point geometry, limiting its ability to learn fine-grained features. We encode the point cloud directly into a graph, using the points as the vertices of the graph. The edges of the graph connect neighboring points that fall within a set radius, enabling the transfer of feature information between these points. This graph representation can adapt to the structure of a point cloud without the necessity to regularize it. Furthermore, to minimize the translation variance in the local graph, we incorporate a graph stabilizer mechanism. This mechanism permits points to align their coordinates based on their unique features, improving the overall effectiveness of the network.

\noindent\textbf{Graph Construction.}
Given the point cloud $\mathbf{P}=(p_1,p_2,...,p_N)$, we construct a graph $\mathbf{G} = (\mathbf{P}, \mathbf{E})$ by using $\mathbf{P}$ as the vertices and connecting a point to its neighbors within a fixed radius $r$ using the well-known fixed radius near-neighbors search algorithm~\cite{bentley1977complexity}, where $\mathbf{E}=\{ (p_i,p_j)| \left \| x_i-x_j \right \|_2<r  \}$. Notably, given a cut-off distance, we utilize a cell list to find point pairs with a runtime complexity of $O(cN)$ where c is the max number of neighbors within the radius, which means this process will not increase much computation complexity.

\noindent\textbf{Graph Stabilizer Mechanism.}
Typically, we can refine the vertex features by aggregating features along the edges within a graph neural network. In our scenario, we aim to include a vertex's local information about the object where the vertex belongs. So in the $(t+1)$-th iteration, we use the relative coordinates of the neighbors for the edge feature extraction, which can be  denoted as:
\begin{equation}
\label{eq::8}
v_i^{t+1} = g^t( \varrho(f^t(x_j-x_i,v_j^t)| (i,j) \in \mathbf{E} ), v_i^t ), 
\end{equation}
where $v^t$ is the vertex feature from the $t$-th iteration, $f^t(\cdot)$ computes the edge feature between two vertices, $g^t(\cdot)$ updates the vertex features, and $ \varrho(\cdot)$ is a set function that is used to accumulate the edge features for each vertex.

Despite its effectiveness, it remains susceptible to translation within the local vicinity. A slight translation added to a vertex does not significantly alter the local structure of its neighboring vertices. However, it modifies the relative coordinates of these neighbors, thereby escalating the input variance to function $f^t(\cdot)$. To diminish this translation variance, we propose the alignment of neighboring coordinates based on their structural features rather than relying on the central vertex coordinates. Since the central vertex already contains some structural features from the previous iteration, it can be used to estimate an alignment offset, prompting us to design a graph stabilizer mechanism. The Equation~\ref{eq::8} can be rewritten as:
\begin{equation}
\begin{aligned}
&  \Delta x_i^t = h^t(v_i^t)  \\
& v_i^{t+1} = g^t( \varrho(f^t(x_j-x_i+\Delta x_i^t ,v_j^t)| (i,j) \in \mathbf{E} ), v_i^t ), 
\end{aligned}
\end{equation}
where $h^t(\cdot)$ calculates the offset using the center vertex feature value from the last iteration and 
$\Delta x_i^t$ is the coordination offset for the vertices to stabilize their coordinates. As shown in Figure~\ref{fig::pointrwkv}(b), we model $f^t(\cdot)$, $g^t(\cdot)$ and $h^t(\cdot)$ by multi-layer perception (MLP) and add a residual connection in $g^t(\cdot)$.
\section{Experiments}
\label{experiments}
To validate the effectiveness of our proposed PointRWKV, we conduct the following experiments: a) pre-training our model on ShapeNet~\cite{chang2015shapenet} training set, b) evaluating our pre-trained model on various downstream tasks, including 3D object classification, 3D part segmentation, and few-shot learning, c) ablating on various strategies. Furthermore, we also conduct the experiments that training from scratch. These results can be found in the \textbf{supplementary material}.  

\begin{table*}[t]
  \centering
  \resizebox{0.98\linewidth}{!}{
    \begin{tabular}{cccccccc}
    \toprule
    \multirow{2}[2]{*}{Method} & \multicolumn{3}{c}{ScanObjectNN} & \multicolumn{2}{c}{ModelNet40} & \multirow{2}[2]{*}{Params(M) $\downarrow$} & \multirow{2}[2]{*}{FLOPs(G) $\downarrow$} \\
    \cmidrule(r){2-4} \cmidrule(l){5-6}
       & OBJ-BG $\uparrow$ & OBJ-ONLY $\uparrow$ & PB-T50-RS $\uparrow$ & OA (\%) $\uparrow$  & mAcc (\%) $\uparrow$ &   &  \\
    \midrule
     \multicolumn{8}{c}{\emph{MLP/CNN-based}} \\
    \midrule
      PointNet~\cite{qi2017pointnet}  & 73.3  & 79.2  & 68.0  & 89.2  & 86.2  & 3.5  &  0.5 \\
      PointNet++~\cite{qi2017pointnet++}    & 82.3 & 84.3 & 77.9  & 90.7  & -  & 1.5  & 1.7 \\
      PointCNN~\cite{li2018pointcnn} & 86.1 & 85.5 & 78.5  & 92.2 & 88.1 & 0.6 & 0.9 \\
      DGCNN~\cite{wang2019dynamic} &  82.8& 86.2 & 78.1 & 92.9 & - & 1.8 & 2.4 \\
      MVTN~\cite{hamdi2021mvtn} & 92.6& 92.3 & 82.8  & 93.8 & - & 11.2 & 43.7 \\
      PointMLP~\cite{ma2022rethinking} & - & - &  85.4±0.3  &  94.5 & 91.4 & 12.6 & 31.4 \\
      PointNeXt~\cite{qian2022pointnext} & - & -&  87.7±0.4 & 93.2±0.1 & 90.8±0.2 & 1.4 & 1.6 \\
      \midrule
     \multicolumn{8}{c}{\emph{Transformer-based}} \\
    \midrule
    Transformer~\cite{vaswani2017attention}  & 79.86 & 80.55 & 77.24 & 91.4 & - & 22.1 & 4.8 \\
    PointTransformer~\cite{zhao2021point}  & - & - & -& 93.7 &90.6 & 22.1 & 4.8 \\
    Point-BERT~\cite{yu2022point}  & 87.43 & 88.12 & 83.07& 93.2 &- & 22.1 & 4.8 \\
    Point-MAE~\cite{pang2022masked} &90.02 & 88.29 & 85.18& 93.8 &- & 22.1 & 4.8 \\
    Point-M2AE~\cite{zhang2022point} & 91.22 & 88.81 & 86.43 & 94.0 &- & 15.3 & 3.6 \\
    \midrule
    \multicolumn{8}{c}{\emph{Mamba-based}} \\
    \midrule
    PCM~\cite{zhang2024point} & - & - & 88.10±0.3 & 93.4±0.2& - & 34.2 & 45.0 \\
    PointMamba~\cite{liang2024pointmamba}  & 90.71 & 88.47 & 84.87 & 93.6& - & 12.3 & 3.6 \\
    Mamba3D~\cite{han2024mamba3d}  &95.18 & 92.60 & 88.97 &95.1 & - & 16.9 & 3.9 \\
    \midrule
    \multicolumn{8}{c}{\emph{RWKV-based}} \\
    \midrule
    \rowcolor{gray!12}\textbf{PointRWKV w/o vot.}  & \textbf{96.01}  & \textbf{95.62} & \textbf{93.05}  & \textbf{96.16} & \textbf{92.25} & \textbf{10.6} & \textbf{2.1} \\
    \rowcolor{gray!12}\textbf{PointRWKV  w/ vot.}  & \textbf{97.52}  & \textbf{96.58} & \textbf{93.63}  & \textbf{96.89} & \textbf{93.08} & \textbf{10.6} & \textbf{2.1} \\
    \bottomrule
    \end{tabular}
    }
    \caption{\textbf{Object classification on the ScanObjectNN and ModelNet40 datasets.} We report the overall accuracy (OA, \%) of ScanObjectNN on its three variants: OBJ-BG, OBJ-ONLY, and PB-T50-RS, the OA and mAcc of ModelNet40 and the overall params(M) and FLOPs(G).}
  \label{tab:clsresults}
  \vspace{-0.45cm}
\end{table*}
\subsection{Implementation Details}
\label{datasetsmetrics}
Following Point-MAE~\cite{pang2022masked}, we utilized the ShapeNet~\cite{chang2015shapenet} dataset for pre-training. The configurations include using the AdamW optimizer~\cite{loshchilov2017decoupled} with a learning rate of 1e-3, a weight decay of 5e-2, and a cosine learning rate scheduler. The training consists of 300 epochs with a warm-up period of 10 epochs, using a batch size of 128. The model architecture comprises 12 encoder layers and 4 decoder layers. For multi-scale point cloud learning, we set the scale number $M$ to 3. For the 3-scale point clouds, we set the point numbers as $\{2048, 1024, 512\}$ with a masking ratio of 80$\%$. We also set different $k$ for the $k$-NN at different scales, which are $\{16, 8, 8\}$. In LGM, We use three iterations in our constructed graph. Data augmentation techniques involving scaling and translation are also applied. 
When fine-tuning the model for downstream tasks, such as classification and segmentation, we maintain the use of the AdamW optimizer but adjust the learning rate to 3e-4 for ModelNet40~\cite{wu20153d} and ScanObjectNN~\cite{uy2019revisiting} classification tasks, and to 5e-4 for the ShapeNetPart~\cite{yi2016scalable} segmentation task, with a weight decay of 5e-2 and a cosine scheduler. The number of training epochs remains at 300, with a batch size adjusted to 32 for classification tasks and 16 for the segmentation task. The model architecture is consistent with the pre-training setup, with the same number of encoder and decoder layers, and input points. The augmentation strategy is adapted to fit each specific task, ensuring the model's robustness and generalizability. 

\subsection{Comparison on Downstream Tasks}

\begin{table}[ht]
\centering
\begin{adjustbox}{width=\linewidth}
	\begin{tabular}{lc c c c c}
	\toprule
		\makecell*[c]{\multirow{2}*{Method}} &\multicolumn{2}{c}{5-way} &\multicolumn{2}{c}{10-way}\\
		 \cmidrule(lr){2-3} \cmidrule(lr){4-5}
		 &10-shot $\uparrow$ &20-shot $\uparrow$ &10-shot $\uparrow$ &20-shot $\uparrow$\\
		 \cmidrule(lr){1-1} \cmidrule(lr){2-5}
		DGCNN~\cite{wang2019dynamic} &91.8\ $\pm$\ 3.7 &93.4\ $\pm$\ 3.2 &86.3\ $\pm$\ 6.2 &90.9\ $\pm$\ 5.1\\
		  DGCNN + OcCo~\cite{wang2021unsupervised} &91.9\ $\pm$\ 3.3 &93.9\ $\pm$\ 3.1 &86.4\ $\pm$\ 5.4 &91.3\ $\pm$\ 4.6\\
		\cmidrule(lr){1-5}
	    Transformer~\cite{yu2022point} &87.8\ $\pm$\ 5.2 &93.3\ $\pm$\ 4.3 &84.6\ $\pm$\ 5.5 &89.4\ $\pm$\ 6.3\\
		 Transformer + OcCo~\cite{yu2022point} &94.0\ $\pm$\ 3.6 &95.9\ $\pm$\ 2.3 &89.4\ $\pm$\ 5.1 &92.4\ $\pm$\ 4.6\\
		 Point-BERT~\cite{yu2022point} &94.6\ $\pm$\ 3.1 &96.3\ $\pm$\ 2.7 &91.0\ $\pm$\ 5.4 &92.7\ $\pm$\ 5.1\\
         Point-M2AE ~\cite{zhang2022point} &96.8\ $\pm$\ 1.8& 98.3\ $\pm$\ 1.4&92.3\ $\pm$\ 4.5&95.0\ $\pm$\ 3.0\\
        \midrule
        PointMamba~\cite{liang2024pointmamba} & 95.0\ $\pm$\ 2.3 & 97.3\ $\pm$\ 1.8 & 91.4\ $\pm$\ 4.4 & 92.8\ $\pm$\ 4.0 \\
        Mamba3D~\cite{han2024mamba3d}  &96.4\ $\pm$\ 2.2 & 98.2\ $\pm$\ 1.2 &92.4\ $\pm$\ 4.1 & 95.2\ $\pm$\ 2.9 \\
        \midrule
	     \rowcolor{gray!12}\textbf{PointRWKV(ours)} &\textbf{97.9\ $\pm$\ 1.2}& \textbf{99.2\ $\pm$\ 0.6}&\textbf{94.8\ $\pm$\ 2.8} &\textbf{96.7\ $\pm$\ 2.6}\\
	    \textit{Improvement} &\textcolor{gray}{+1.1} &\textcolor{gray}{+0.9} &\textcolor{gray}{+2.5} &\textcolor{gray}{+1.5}\\
	\bottomrule
	\end{tabular}
\end{adjustbox}
\caption{\textbf{Few-shot classification on ModelNet40}. We report the average accuracy ($\%$) and standard deviation ($\%$) of 10 independent experiments without voting. 
}
\label{tab:fewshot}
\end{table}

\begin{table*}[t]
  \centering
  \resizebox{0.95\linewidth}{!}{
    \begin{tabular}{cccccc}
    \toprule
    Method & Backbone & Cls. mIoU (\%) $\uparrow$ & Inst. mIoU (\%) $\uparrow$ & Params(M) $\downarrow$ & FLOPs(G) $\downarrow$ \\
    \midrule
      PointNet~\cite{qi2017pointnet}   & MLP-based & 80.39 & 83.7 & 3.6  & 4.9 \\
      PointNet++~\cite{qi2017pointnet++}   & MLP-based & 81.85  & 85.1  & 1.0 & 4.9 \\
      PointCNN~\cite{li2018pointcnn} & CNN-based & 84.6 & 86.1 & - & - \\
      DGCNN~\cite{wang2019dynamic}  & CNN-based & 82.3 & 85.2 & 1.3 & 12.4 \\
      APES~\cite{wu2023attention}  & MLP-based  & 83.67 & 85.8 & - & - \\
      PointMLP~\cite{ma2022rethinking} &  MLP-based  &  84.6 & 86.1 & - & - \\
      PointNeXt~\cite{qian2022pointnext} & MLP-based &  85.2±0.1 & 87.0±0.1 & 22.5 & 110.2 \\
      \midrule
    Transformer~\cite{vaswani2017attention}  & Transformer-based & 83.4 & 85.1 & 27.1 & 15.5\\
    PointTransformer~\cite{zhao2021point}   & Transformer-based & 83.7 &86.6 & - & - \\
    Point-BERT~\cite{yu2022point}  & Transformer-based & 84.1 &85.6 &27.1 & 10.6 \\
    Point-MAE~\cite{pang2022masked}  & Transformer-based & 84.2 &86.1 & 27.1 & 15.5 \\
    Point-M2AE~\cite{zhang2022point}& Transformer-based & 84.86 &86.51 & - &- \\
    \midrule
    PCM~\cite{zhang2024point}  & Mamba-based & 85.6 & 87.1 & 34.2 & 45.0 \\
    PointMamba~\cite{liang2024pointmamba}  & Mamba-based & 84.42 & 86.0 & 17.4 & 14.3\\
    Mamba3D~\cite{han2024mamba3d}  & Mamba-based &84.1 & 85.7  & 21.9  & 9.5 \\
    \midrule
   \rowcolor{gray!12} \textbf{PointRWKV(ours)} & RWKV-based  & \textbf{89.87} & \textbf{90.26} & \textbf{15.2} & \textbf{6.5} \\
    \bottomrule
    \end{tabular}
    }
  \caption{\textbf{Part segmentation on ShapeNetPart dataset.} The class mIoU and the instance mIoU are reported, with model parameters (M) and FLOPs (G).}
  \label{tab:partresults}
  \vspace{-0.25cm}
\end{table*}
\noindent
\textbf{Real-World Object Classification on ScanObjectNN Dataset.}
ScanObjectNN~\cite{uy2019revisiting}, a highly challenging 3D dataset, encompasses approximately 15,000 objects distributed across 15 distinct categories. These objects, derived from real-world indoor environments characterized by cluttered backgrounds, contribute to the heightened complexity inherent in the task of object classification. As shown in Table~\ref{tab:clsresults}, we conduct experiments on its three variants: OBJ-BG, OBJ-ONLY, and PB-T50-RS. After pre-training, our proposed PointRWKV consistently outperforms transformer- and mamba-based models.  Specifically, PointRWKV outperforms previous sota Mamba3D by 2.34$\%$, 3.98$\%$ and 4.66$\%$ on three variants while reducing 37.2$\%$ parameters and 46.2$\%$ FLOPs. When compared to the transformer-based methods like Point-BERT, Point-MAE and Point-M2AE, PointRWKV demonstrates a substantial improvement, achieving 10.56$\%$, 8.45$\%$ and 7.2$\%$ higher accuracy on the challenging PB-T50-RS subset, respectively. Even without the voting strategy~\cite{liu2019relation}, our PointRWKV still performs better than other methods that are tested with voting strategy.

\noindent
\textbf{Synthetic Object Classification on ModelNet40 Dataset.} 
ModelNet40~\cite{wu20153d} is a CAD 3D object classification dataset, which includes over 12K synthetic 3D CAD models cover 40 categories in total.
As shown in Table~\ref{tab:clsresults}, we report overall accuracy (OA) and mAcc results. PointRWKV maintains its dominance with an OA of 96.89$\%$, which is 1.79$\%$ higher than Mamba3D's 95.1$\%$. PointRWKV also shows a 2.89$\%$ improvement over Point-M2AE, which has an OA of 94.0$\%$. Similarly, our PointRWKV still outperforms other methods even without the voting strategy~\cite{liu2019relation}. 
Overall, these results highlight PointRWKV’s superiority over existing transformer- or mamba-based models, achieving multiple SoTA and demonstrating its strength across various settings.

\begin{table*}
\begin{minipage}[t!]{0.32\linewidth}
\centering
 \tiny
\begin{adjustbox}{width=\linewidth}
\centering
	\begin{tabular}{lcl}
	\toprule
		Variant &OA  & mAcc \\
		\cmidrule(lr){1-1} \cmidrule(lr){2-2} \cmidrule(lr){3-3}
      base &89.6 &88.3 \\
      + BQE &90.4 &89.0\\
      + bi-attention &91.6 &90.2 \\
      + Multi-scale &92.8 &90.6  \\
      + LGM w/o GS &93.0 &91.1 \\
      + LGM w/ GS &93.6 &91.7 \\ 
	  \bottomrule
	\end{tabular}
\end{adjustbox}
\caption{\textbf{Ablation of main components}.}
\label{t4}
\end{minipage}
\hspace{0.2cm}
\begin{minipage}[t!]{0.33\linewidth}
\centering
 \tiny
\begin{adjustbox}{width=\linewidth}
\centering
	\begin{tabular}{ccc}
	\toprule
		Scales &OA &mAcc \\
		\cmidrule(lr){1-1} \cmidrule(lr){2-2} \cmidrule(lr){3-3} 
      512 &90.3 &88.9  \\
      1024  &89.6 &88.3\\
      2048  &90.2 &89.5 \\
      4096 &89.7 & 88.8 \\
      512-1024-2048 &92.8 &91.6\\
     1024-2048-4096 &91.5 &90.1 \\
	  \bottomrule
	\end{tabular}
\end{adjustbox}
\caption{\textbf{Ablation of different point scales}.}
\label{t5}
\end{minipage}
\hspace{0.2cm}
\begin{minipage}[t!]{0.3\linewidth}
\centering
 \tiny
\begin{adjustbox}{width=\linewidth}
\centering
	\begin{tabular}{clc}
	\toprule
      Number &OA &mAcc \\
		\cmidrule(lr){1-1} \cmidrule(lr){2-2} \cmidrule(lr){3-3} 
      0 &91.2 &90.5  \\
      1 &92.1	&91.2  \\
      2 &92.4	&91.4  \\
      3 &92.5 &91.7\\
      4 &92.3	&91.5 \\
	  \bottomrule
	\end{tabular}
\end{adjustbox}
\caption{\textbf{Ablation of different number of graph iteration}.}
\label{t6}
\end{minipage}
\end{table*}

\begin{figure}[t]
    \centering
\includegraphics[width=1\linewidth]{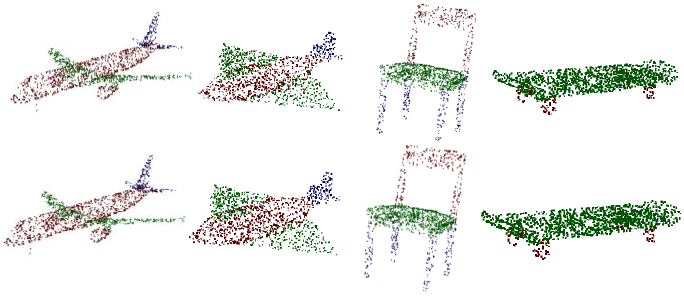}
    \caption{\textbf{Part segmentation results on ShapeNetPart.}  Top
row is ground truth and bottom row is our prediction.}
    \label{fig::vis}
\end{figure}

\noindent
\textbf{Part Segmentation on ShapeNetPart Dataset.}
We conduct part segmentation on the challenging ShapeNetPart~\cite{yi2016scalable} dataset to predict more detailed class labels for each point within a sample. It comprises 16880 models with 16 different shape categories and 50 part labels. Experimental results on the ShapeNetPart dataset are shown in Table~\ref{tab:partresults}. We report mean IoU (mIoU) for all classes (Cls.) and all instances (Inst.). Our PointRWKV still obtains 3.16$\%$ and 4.27$\%$ improvements in terms of Inst. mIoU and Cls. mIoU while significantly reducing parameters and FLOPs. Moreover, we present qualitative results of part segmentation in Figure~\ref{fig::vis}. It can be found that PointRWKV achieves highly competitive results compared to the ground truth.

\noindent
\textbf{Few-shot Learning.}
To further evaluate the performance of PointRWKV with limited fine-tuning data, we conduct experiments for few-shot classification on ModelNet40 with an $n$-way, $m$-shot setting, where $n$ is the number of classes randomly sampled from the dataset, and $m$ denotes the number of samples randomly drawn from each class. As shown in Table~\ref{tab:fewshot}, we experiment with $n \in \{ 5, 10\}$ and $m \in \{ 10, 20\}$. PointRWKV surpasses previous methods by 1.1$\%$, 0.9$\%$, 2.5$\%$, and 1.5$\%$, respectively, for all four settings with much smaller deviations. This illustrates PointRWKV's competence in learning semantic information and its potent ability to transfer knowledge to downstream tasks, even with limited data.

\noindent
\textbf{Structural Efficiency.} 
To fully explore the the efficiency of our method, we gradually increase the sequence length to test the ability of processing the long point sequence. As shown as Figure~\ref{fig:acc_time}, when the sequence continues to increase, we reduce the GFLOPs by 15.4$\times$ and 2.1$\times$ compared with the transformer-based method and the most convincing mamba-based method, achieving true linear complexity. Furthermore, as illustrated in the last two columns of Table~\ref{tab:clsresults} and Table~\ref{tab:partresults}, our PointRWKV reduces 3.1$\%$-13.8$\%$ in parameters and 31.6$\%$-41.7$\%$ in FLOPs when compared with transformer- and mamba-based counterparts.

\subsection{Ablation Study}
\label{ablations}

To investigate the effectiveness of each component and design, we conduct ablation studies on ScanObjectNN with the PB-T50-RS variant unless specified.

\noindent
\textbf{Analysis of Different Components.}
We illustrate the importance of different components of our network by adding different parts. The baseline setting is to only use single scale of 1024 points as input and remove the BQE, bidirectional attention mechanism, the LGM and graph stabilizer (GS) mechanism. As shown in Table~\ref{t4}, we add the shift by bidirectional quadratic expansion (BQE) function and the modified bidirectional attention mechanism, enhancing OA almost equally, respectively. We also observe that applying the hierarchical multi-scale point cloud learning and local graph-based merging can lift the performance by a large margin, which demonstrates the importance of refined feature learning. And the absence of graph stabilizer hurts the performance, which also proves the necessity of local graph adjustments.

\noindent
\textbf{Ablations of Different Point Scales.}
In Table~\ref{t5}, we experiment with different scales of points as input to encode the token embeddings. Apart from the original single 1024 points input, we also sample different points and combine adjacent scale to form the multi-scales embeddings. It can be observed that single-scale input fluctuates in accuracy, but multi-scale input has more outstanding final performance.

\noindent
\textbf{Ablations of Different Number of Graph Iteration.}
In local graph-based merging branch,we refine the vertex states iteratively. In Table~\ref{t6}, we study the impact of the number of iterations on the accuracy. The initial vertex state alone achieves the lowest accuracy since it only has a small receptive field around the vertex. As the number of iterations increases, the corresponding accuracy value also becomes higher compared to no iteration. This largely illustrates the effectiveness of the iterative graph stabilizer.

\begin{table}[htbp]
  \centering
  \resizebox{0.55\linewidth}{!}{
  \tiny
    \begin{tabular}{ccc}
	\toprule
		Ratio &OA &mAcc \\
		\cmidrule(lr){1-1} \cmidrule(lr){2-2} \cmidrule(lr){3-3} 
        0 &   87.3  & 86.2  \\
      0.5 &91.5 &90.6  \\
      0.6  &91.8 &90.8\\
      0.7  &92.4 &91.3 \\
      0.8 &92.8 & 91.6 \\
      0.9 &92.1 &91.2\\
	  \bottomrule
	\end{tabular}
    }
  \caption{\textbf{Different Masking Strategy.} }
  \label{tab:mask}
\end{table}

\noindent
\textbf{Ablation of Different Masking Strategy.}
In Table~\ref{tab:mask}, we report the results of different masking strategies during the multi-scale points processing. For different mask ratios, we find the 80$\%$ ratio performs the best to build a proper representation for hierarchical point cloud learning.

\section{Conclusion}
In this paper, we introduce PointRWKV, a novel RWKV-based architecture for point cloud learning. With a hierarchical architecture, PointRWKV learns to produce powerful 3D representations by encoding multi-scale point clouds. To facilitate local and global feature aggregation, we design the parallel feature merging strategy. 
Experimental results show that PointRWKV exhibits superior performance over transformer- and mamba-based counterparts on different point cloud learning datasets while significantly reducing parameters and FLOPs. 
With its linear complexity capability, we hope PointRWKV will serve as an efficient and cost-effective baseline for more 3D tasks.

\noindent
\textbf{Limitations and Broader Impact.} The model still relies on the pre-training scheme, and more elegant methods should be designed. This study makes the initial attempt to apply RWKV in 3D point cloud learning, laying a foundation for future research.

{
\small
\bibliographystyle{ieeenat_fullname}
\bibliography{main}
}

\clearpage
\renewcommand{\thefigure}{A\arabic{figure}}
\setcounter{figure}{0}
\renewcommand{\thetable}{A\arabic{table}}
\setcounter{table}{0}
\renewcommand{\thesection}{A\arabic{section}}
\setcounter{section}{0}
\maketitlesupplementary

The supplementary material presents the following sections to strengthen the main manuscript:

\begin{itemize}
\item Results of training from scratch.
\item Semantic segmentation on S3DIS dataset.
\item More qualitative visualization results on ShapeNetPart.
\end{itemize}

\section{Results of Training From Scratch}
To further demonstrate the effectineness of our proposed PointRWKV, we also report the results for models trained from scratch. It is noted that compared with pre-training, many methods does not report the results of training from scratch. As shown in Table ~\ref{tab:trainfromscratchclsresults}, we conduct experiments on ScanObjectNN and ModelNet40 Dataset. Compared to previous methods, PointRWKV obtains 0.51$\%$-2.9$\%$ overall accuracy performance gains across different variants among different datasets. When without voting strategy, PointRWKV can still outperform other methods.

\begin{table*}[t]
  \centering
  \resizebox{0.98\linewidth}{!}{
    \begin{tabular}{cccccccc}
    \toprule
    \multirow{2}[2]{*}{Method} & \multicolumn{3}{c}{ScanObjectNN} & \multicolumn{2}{c}{ModelNet40} & \multirow{2}[2]{*}{Params(M) $\downarrow$} & \multirow{2}[2]{*}{FLOPs(G) $\downarrow$} \\
    \cmidrule(r){2-4} \cmidrule(l){5-6}
       & OBJ-BG $\uparrow$ & OBJ-ONLY $\uparrow$ & PB-T50-RS $\uparrow$ & OA (\%) $\uparrow$  & mAcc (\%) $\uparrow$ &   &  \\
    \midrule
     \multicolumn{8}{c}{\emph{MLP/CNN-based}} \\
    \midrule
      PointNet~\cite{qi2017pointnet}  & 73.3  & 79.2  & 68.0  & 89.2  & 86.2  & 3.5  &  0.5 \\
      PointNet++~\cite{qi2017pointnet++}    & 82.3 & 84.3 & 77.9  & 90.7  & -  & 1.5  & 1.7 \\
      PointCNN~\cite{li2018pointcnn} & 86.1 & 85.5 & 78.5  & 92.2 & 88.1 & 0.6 & 0.9 \\
      DGCNN~\cite{wang2019dynamic} &  82.8& 86.2 & 78.1 & 92.9 & - & 1.8 & 2.4 \\
      MVTN~\cite{hamdi2021mvtn} & 92.6& 92.3 & 82.8  & 93.8 & - & 11.2 & 43.7 \\
      PointMLP~\cite{ma2022rethinking} & - & - &  85.4±0.3  &  94.5 & 91.4 & 12.6 & 31.4 \\
      PointNeXt~\cite{qian2022pointnext} & - & -&  87.7±0.4 & 93.2±0.1 & 90.8±0.2 & 1.4 & 1.6 \\
      \midrule
     \multicolumn{8}{c}{\emph{Transformer-based}} \\
    \midrule
    PointTransformer~\cite{zhao2021point}  & - & - & -& 93.7 &90.6 & 22.1 & 4.8 \\
    \midrule
    \multicolumn{8}{c}{\emph{Mamba-based}} \\
    \midrule
    PCM~\cite{zhang2024point} & - & - & 88.10±0.3 & 93.4±0.2& - & 34.2 & 45.0 \\
    PointMamba~\cite{liang2024pointmamba}  & 88.30 & 87.78 & 82.48 & 92.4& - & 12.3 & 3.6 \\
    Mamba3D~\cite{han2024mamba3d}  &94.49 & 92.43 & 92.64 &94.1 & - & 16.9 & 3.9 \\
    \midrule
    \multicolumn{8}{c}{\emph{RWKV-based}} \\
    \midrule
    \rowcolor{gray!12}\textbf{PointRWKV w/o vot.}  & \textbf{95.22}  & \textbf{94.28} & \textbf{92.88}  & \textbf{94.66} & \textbf{91.35} & \textbf{10.6} & \textbf{2.1} \\
    \rowcolor{gray!12}\textbf{PointRWKV  w/ vot.}  & \textbf{96.53}  & \textbf{95.33} & \textbf{93.15}  & \textbf{95.36} & \textbf{92.48} & \textbf{10.6} & \textbf{2.1} \\
    \bottomrule
    \end{tabular}
    }
    \caption{\textbf{Object classification on the ScanObjectNN and ModelNet40 datasets of training from scratch.} We report the overall accuracy (OA, \%) of ScanObjectNN on its three variants: OBJ-BG, OBJ-ONLY, and PB-T50-RS, the OA and mAcc of ModelNet40 and the overall params(M) and FLOPs(G).}
  \label{tab:trainfromscratchclsresults}
\end{table*}

\section{Semantic Segmentation on S3DIS Dataset}
S3DIS~\cite{armeni20163d} dataset consists of 271 rooms in six areas from three different buildings. Each point in the scan is assigned a semantic label from 13 categories. we evaluate on Area 5 as shown in Table ~\ref{tab:semantic}.  The results further demonstrate the superiority of our method over transformer- and mamba-based counterpart.

\begin{table*}[htbp]
  \centering
  \resizebox{\linewidth}{!}{
    \begin{tabular}{c|ccc|ccccccccccccc}
	\toprule
		Method &OA &mAcc&mIoU &ceiling &floor &wall &beam &column &window &door &table &chair &sofa &bookcase &board &clutter \\
		\cmidrule(lr){1-17} 
      PointNet~\cite{qi2017pointnet} &- &49.0& 41.1& 88.8& 97.3& 69.8& 0.1& 3.9& 46.3& 10.8& 59.0& 52.6& 5.9& 40.3& 26.4& 33.2\\
      PointCNN~\cite{li2018pointcnn} &85.9& 63.9& 57.3& 92.3& 98.2& 79.4& 0.0& 17.6& 22.8& 62.1& 74.4 &80.6 &31.7 &66.7 &62.1 &56.7\\
      PointNeXt-S~\cite{qian2022pointnext}& 87.9& -& 63.4& - &-& -& -& -& -& -& - &- &- &-& - &- \\
      PointCloudMamba~\cite{zhang2024point}&88.2& 71.0 &63.4& 93.3 &96.7& 80.6& 0.1 &35.9 &57.7 &60.0& 74.0& 87.6& 50.1& 69.4& 63.5& 55.9\\
      \midrule
      \textbf{PointRWKV(ours)} & \textbf{92.3} & \textbf{76.8} & \textbf{70.5} & 94.2 & 98.3& 86.5& 0.0& 38.6& 64.5& 76.2& 88.2 & 89.3& 65.2& 75.6& 78.2& 61.3\\
	  \bottomrule
	\end{tabular}
    }
  \caption{\textbf{Semantic segmentation results on S3DIS dataset.} }
  \label{tab:semantic}
\end{table*}

\section{More Qualitative Visualization on ShapeNetPart}
\begin{figure}[ht]
\centering
\includegraphics[width=0.65\linewidth]{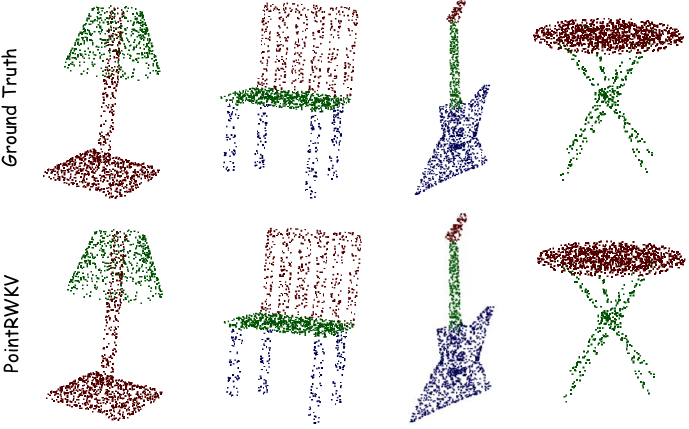}
\caption{\textbf{Qualitative results of part segmentation results on ShapeNetPart.}  Top
row is ground truth and bottom row is our prediction.}
\label{fig::supplyvis}
\end{figure}

In Figure~\ref{fig::supplyvis}, we illusttrate more Visualizations which further shows that PointRWKV achieves highly competitive results to the  ground truth.
 
\clearpage
\cleardoublepage


\end{document}